\documentclass[conference]{IEEEtran}
\IEEEoverridecommandlockouts
\usepackage{cite}
\usepackage{amsmath,amssymb,amsfonts}
\usepackage{graphicx}
\usepackage{xcolor}
\usepackage{comment, lipsum}
\usepackage{amsmath,amssymb} 
\usepackage{tikz}
\usepackage{color}
\usepackage{multirow, multicol}
\usepackage{graphicx, caption, subcaption}
\usepackage{newtxtext,newtxmath}
\usepackage{subcaption}
\usepackage{newtxmath}
\usepackage{algorithm}
\usepackage{algpseudocode}
\usepackage{varwidth}
\usepackage{tabulary,booktabs}

\begin{document}

\title{Visual Heart Rate Estimation from RGB Facial Video using Spectral Reflectance
}

\author{
\IEEEauthorblockN{Bharath Ramakrishnan}
\IEEEauthorblockA{
\textit{Texas A\&M University}\\
Texas, USA}
\and
\IEEEauthorblockN{Ruijia Deng}
\IEEEauthorblockA{
\textit{Wuhan University} \\
Wuhan, China}
\and
\IEEEauthorblockN{Hassan Ali}
\IEEEauthorblockA{
\textit{Texas A\&M University}\\
Texas, USA}
}

\maketitle

\begin{abstract}
Estimation of the Heart rate from facial video has a number of applications in the medical and the fitness industries. Additionally, it has become useful in the field of gaming as well. Several approaches have been proposed to seamlessly obtain the Heart rate from the facial video, but these approaches have had issues in dealing with motion and illumination artifacts. In this work, we propose a reliable HR estimation framework using the spectral reflectance of the user, which makes it robust to motion and illumination disturbances. We employ deep learning-based frameworks such as Faster RCNNs to perform face detection as opposed to the Viola Jones algorithm employed by previous approaches. We evaluate our method on the MAHNOB-HCI dataset and found that the proposed method is able to outperform previous approaches. 
\end{abstract}

\section{Introduction}

The estimation of medical parameters such as Heart Rate, Heart Rate variability have been widely useful for the determination of fitness and emotion in the sports and medical industry. Additionally, measures such as Heart Rate variability have also been used as a method to detect depression in patients. Using conventional techniques such as polar Heart Rate sensors can often be intrusive for the patient and prevent long-term monitoring of the Heart Rate especially for medical conditions. Additionally, they can be uncomfortable for the patient since the pressure can become inconvenient over time. Over the years, it has been shown that it is possible to extract the Heart Rate from facial video without requiring any heart rate sensors, while achieving the accuracy of traditional Heart Rate sensors \cite{balakrishnan2013detecting, cennini2010heart, icaart}.  
There have been several approaches to measuring Heart Rate from an RGB video of the user's face. The two main approaches have been the following:
\begin{itemize}
    \item \textbf{Motion-based Estimation}: Inspired from \cite{balakrishnan2013detecting}, several approaches have documented the use of head pulse motions to measure heart rate. The primary concept of the approach lies in the fact that the head asserts a Newtonian reaction to the pulse of the blood flowing through the user's face as a result of the heart rate. The rate at which the head oscillates is related to the force/rate at which the heart pumps blood through the face. This method, while technically sound, can be problematic in the presence of motion artifacts since it would be difficult to observe the microscopic motions of the user's head due to their heart beat. 
    \item \textbf{Color-based Estimation}: Recently, a majority of approaches have used the remote photoplethysmographic (rPPG) signals from the RGB pixel intensity of the user's face. Its been found that the rPPG signal corresponding to the green channel of the facial video provides the most relevant information about the user's facial video. While a number of approaches have accurately measured the facial video, obtaining accurate HR measurements in the presence of significant motion and illumination interferences has been an issue.   
\end{itemize}

In this work, inspired from \cite{lam2015robust, subramaniam2019spectral}, we use a spectral reflectance based method to remotely obtain the heart rate of the user. We test the accuracy of our approach against polar heart rate measurements as well as previous competitive baselines on the popular publicly available dataset MAHNOB-HCI.

\section{Method}
\section{Face Detection using Faster RCNNs}

Contrary to previous work, we use deep learning to apply face detection on real time video. We employ Faster RCNNs to extract facial features and produce a bounding box, which would serve as a Region of Interest (ROI) for downstream processing \cite{ren2015faster}. In order to ensure that our model is able to produce the bounding box/ROI without latency issues, we prune the model by removing weights lesser than a specified threshold. The methodology of pruning is a one-shot pruning paradigm inspired from previous work \cite{subramaniam2020n2nskip, lee2018snip}. Next, we use the KLT feature tracking algorithm followed by Independent component analysis to get the 
final heart rate of the individual. An important aspect of our work is the tolerance to illumination artifacts. We introduce a spectral reflectance-based step to reduce the impact of illumination artifacts such as the following:
\begin{itemize}
    \item Foreground and the background are lit up by different light sources
    \item The light source of the foreground is changing over time
    \item There are sudden flashes in the image/video, due to which the RGB pixel intensity can be extremely high or extremely low for short periods of time.
\end{itemize}

Our illumination Variation Rectification step proposes to resolve such afore-mentioned illumination variations.

\subsection{Illumination Variation Rectification}
We use the spectral reflectance of the user's face to obtain the noise in the pixel intensity due to the user's blood flow. In other words, we use the noise in the pixel intensity of the user's skin pigmentation to neutralize the noise in the pixel intensity of the user's blood flow intensity. This would give us the pure noise-free intensity corresponding to the blood flow of the user. 

\begin{equation}
    I(\rho, t) = I_{b(impure)}(\rho, t) + I_{b(pure)}(\rho, t)\\
\end{equation}
where I is the intensity given by:
\begin{equation}
    I_{IV} = N\cdot I_{impure}(\rho, t) \\
\end{equation}

where $N$ is the approximation constant for measuring the noise in the blood flow intensity. This can be measured by using the Normalized least mean squares approach in adaptive filtering.

\section{Experimental Results}

\subsection{Experimental Setup}
We train the Faster RCNN on face detection datasets by freezing the first few layers, since the first few layers capture lower level features and can be transferred for the facial detection task. Next, we use the trained Faster RCNN to produce bounding boxes on the facial video of the user by processing each video at 60 FPS. We provide the results of our approach on the MAHNOB -HCI dataset as well as traditional heart rate monitoring systems such as the Polar H10 sensor.  

\subsection{Results and Analysis}

Table \ref{tab:mahnob} shows the results of our dataset as compared to previous methods.

\setlength{\tabcolsep}{3.5pt}
\begin{table}[htbp]
\begin{center}
\begin{tabular}{ccccc}
\hline
Framework & $\mu_{error}$ & RMSE (\%) & \% Absolute Error<5bpm & r \\ 
\hline
\vspace{1mm}
Li2014 & 7.7 & 15 & 68.1 & $0.72^{*}$\\
\vspace{1mm}
Lam2015 & 6.7 & 14.1 & 63.1 & $0.75^{*}$\\
\vspace{1mm}
SAM2016 & 5.5 & 14.2 & 72.1 & $0.75^{*}$\\
\vspace{1mm}
Zha2017 & 5.6 & 14.2 & 78.1 & $0.79^{*}$\\
\vspace{1mm}
Sub2019 & 4.7 & 7.8 & 88.1 & $0.82^{*}$\\
\vspace{1mm}
Ours & 4.1 & 6.6 & 94.1 & $0.86^{*}$\\
\hline
\end{tabular}
\end{center}
\caption{Results of measuring HR from facial video on the MAHNOB-HCI dataset.}
\label{tab:mahnob}
\end{table}

\begin{figure}[htp]
    \centering
    \includegraphics[scale=0.5]{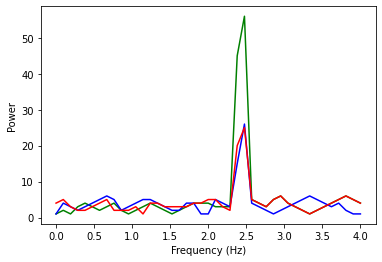}
    \caption{Plot of the heart rate of the user as a power spectra of the rPPG signals corresponding to each RGB channel.}
    \label{fig:HR}
\end{figure}

\begin{figure}[htp]
    \centering
    \includegraphics[scale=0.5]{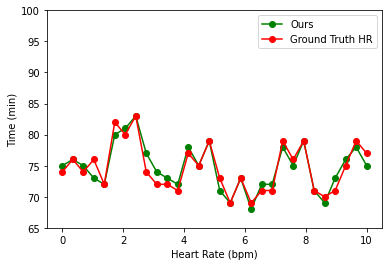}
    \caption{Plot of the heart rate of the user in a 10 min time window and the ground truth plot of the Heart Rate of the user from a Polar H10 Heart Rate sensor}
    \label{fig:HR_ground_truth}
\end{figure}

Finally, we also measure the reliability of our approach against the Polar Heart Rate sensor, which is used as the ground truth Heart Rate measurement. \ref{fig:HR_ground_truth} shows the heart rate measurement of the proposed method as compared to the ground truth HR data over a 10 minute time. We obtain the heart rate of the individual every 10 seconds with a RMSE of only 0.5\%.

\section{Conclusion}
In this work, we propose a method to measure the heart rate of an individual from an RGB video of their face. Our Heart Rate measurement framework is able to obtain an accurate estimate of the user's heart rate even in the presence of considerable motion and illumination artifacts. Additionally, we have moved away fom the Viola Jones Algorithm which is typically used for facial detection by previous works, and instead used a deep learning based framework (Faster RCNN) as a part of our pipeline. Furthermore, we employed an iterative pruning method to reduce the memory footprint of the Faster RCNN so as to reduce the memory footprint and the computational overload on the system hardware.

\bibliography{aaai1}


\end{document}